\documentclass[11pt]{article}
\usepackage{acl}

\usepackage{times}
\usepackage{latexsym}

\usepackage[T1]{fontenc}
\usepackage[utf8]{inputenc}

\usepackage{microtype}

\usepackage{inconsolata}

\usepackage{graphicx}

\usepackage{subcaption}
\usepackage{float}

\title{SyntaxMind at BLP-2025 Task 1: Leveraging Attention Fusion of CNN and GRU for Hate Speech Detection}

\author{Md. Shihab Uddin Riad \\
        Dept. of Computer Science \& Engineering \\ International Islamic University Chittagong \\
        \texttt{shihab.riadn@gmail.com}}

\begin{document}
\maketitle
\begin{abstract}

This paper describes our system used in the BLP-2025 Task 1: Hate Speech Detection. We participated in Subtask 1A and Subtask 1B, addressing hate speech classification in Bangla text. Our approach employs a unified architecture that integrates BanglaBERT embeddings with multiple parallel processing branches based on GRUs and CNNs, followed by attention and dense layers for final classification. The model is designed to capture both contextual semantics and local linguistic cues, enabling robust performance across subtasks. The proposed system demonstrated high competitiveness, obtaining 0.7345 micro F1-Score (2nd place) in Subtask 1A and 0.7317 micro F1-Score (5th place) in Subtask 1B.

\end{abstract}

\section{Introduction}

Hate speech is any type of statement that aims to vilify, humiliate, or instigate hatred toward a group or a class of people based on their nationality, race, religion, skin color, sexual orientation, gender identity, ethnicity, or disability \citep{ward1997free}. With surge of various social media platforms, people often express their opinion without hesitation. However, these opinions are not always appropriate. This study aims to detect hate speech on BLP-2025 Task 1 \citep{blp2025-overview-task1} using the provided multiclass based dataset \citep{hasan2025llm}. BLP-2025 Task 1 consists of three different subtasks. These subtasks focus on detecting hate speech, assessing its severity, and identifying the target group of hate speech in the Bangla language. Our study participates in Subtask 1A and Subtask 1B.

Our work primarily focuses on attention based deep learning techniques to enhance representation and classification performance. As we work through the task, we notice that language specific bias in the pretrained model is significant. Due to that we have utilized BanglaBERT \citep{bhattacharjee-etal-2022-banglabert} as contextual embedding module. Then we pass these embeddings in CNN and Bi-GRU having attention on top classify our data.

\section{Related Work}

Our work is inspired by RiTUAL-UH’s \citep{kar-etal-2017-ritual} approach in SemEval-2017 Task 5, which integrate hand-engineered features with a CNN and a Bi-GRU to predict sentiment scores. Their system utilized Word2Vec for word embeddings and relied significantly on SenticNet to extract critical features.

Similar attention based approaches are evident in the SemEval-2018, NTUA-SLP's \citep{baziotis-etal-2018-ntua} work. The system utilized Bi-LSTM with multiple attention head to classify multi-label emotion in text. They used transfer learning approach by pretraining Bi-LSTMs on the dataset of SemEval-2017, Task 4A to address limited data availability.

\section{Dataset}

\begin{figure*}[ht]
  \centering
  \begin{subfigure}[t]{0.48\linewidth}
    \includegraphics[width=\linewidth]{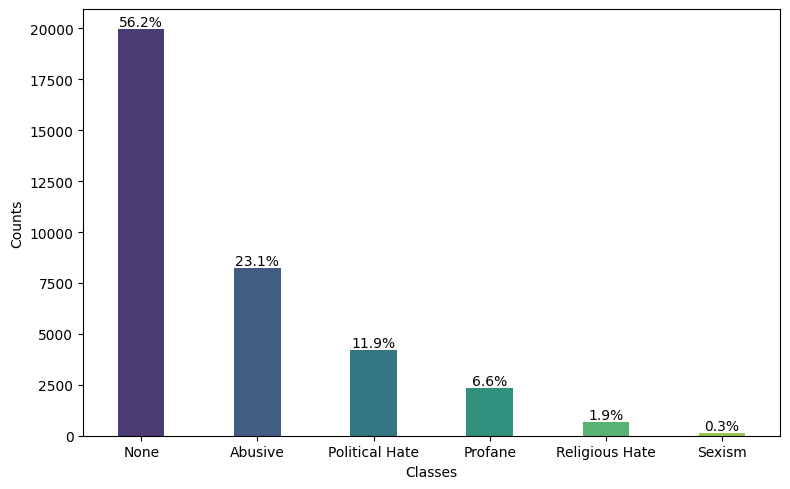}
    \caption{Subtask 1A}
  \end{subfigure}
  \hfill
  \begin{subfigure}[t]{0.48\linewidth}
    \includegraphics[width=\linewidth]{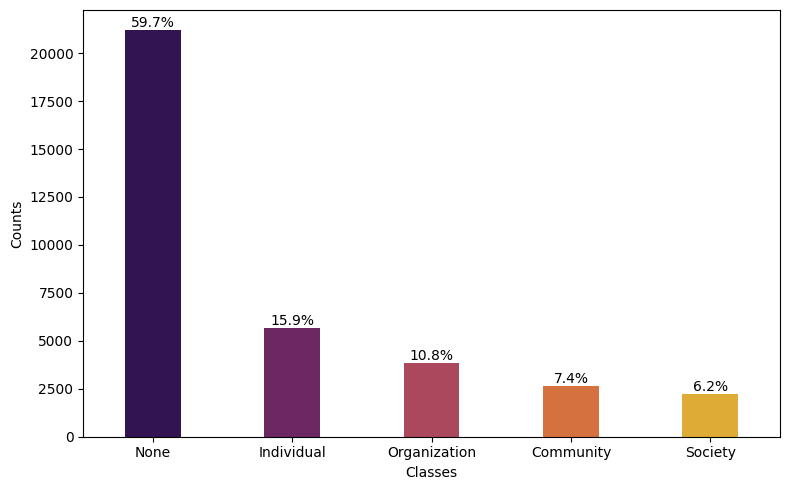}
    \caption{Subtask 1B}
  \end{subfigure}
  \caption{Class distributions for BLP-2025 Task 1}
  \label{fig:class_dist_overall}
\end{figure*}

The datasets for BLP-2025 Task 1, covering Subtask 1A (hate speech detection) and Subtask 1B (target group identification), consist of 35,522 instances each. In Subtask 1A, the "None" (non-hate) category leads with 19,954 samples (56.2\%), trailed by "Abusive" at 8,212 (23.1\%), "Political Hate" at 4,227 (11.9\%), "Profane" at 2,331 (6.6\%), "Religious Hate" at 676 (1.9\%), and "Sexism" at 122 (0.3\%). This distribution suggests a strong inclination toward non-hate or dominant abusive content, which could skew model performance. For Subtask 1B, the "None" (no target) class prevails with 21,190 samples (59.7\%), reflecting a substantial share of non-directed text. The targeted groups include "Individual" at 5,646 (15.9\%), "Organization" at 3,846 (10.8\%), "Community" at 2,635 (7.4\%), and "Society" at 2,205 (6.2\%), indicating a decreasing prevalence from specific to broader entities.

The observed class imbalances present significant challenges for model training and evaluation. As depicted in Figure~\ref{fig:class_dist_overall} for Subtask 1A and Subtask 1B, the minority classes such as "Sexism" and "Religious Hate" in Subtask 1A, and "Society" and "Community" in Subtask 1B constitute a small fraction of the dataset. This under-representation risks poor model generalization for these critical categories, potentially leading to biased predictions that favor the majority "None" class.

\subsection{Preprocessing}

The preprocessing pipeline for the Bangla text dataset is carefully crafted to optimize model performance by addressing linguistic and structural variations. Initially, URLs are removed and English characters are converted to lowercase for uniformity. Emojis are transformed into Bangla text where feasible and numbers separated by commas(,) are merged into a single value. Subsequently, the text undergoes normalization through the Normalization Form Canonical Composition scheme using the unicodedata module, alongside the bnunicodenormalizer\footnote{\url{https://github.com/mnansary/bnUnicodeNormalizer}} \citep{ansary-etal-2024-unicode-normalization}, which decomposes and recomposes characters while eliminating optional zero-width joiners and Bangla punctuation to maintain consistency. Additionally, percentage symbols are substituted with appropriate Bangla term, resulting in a refined and standardized dataset ready for further analysis.

\begin{figure*}[htbp]
    \centering
    \includegraphics[width=\textwidth]{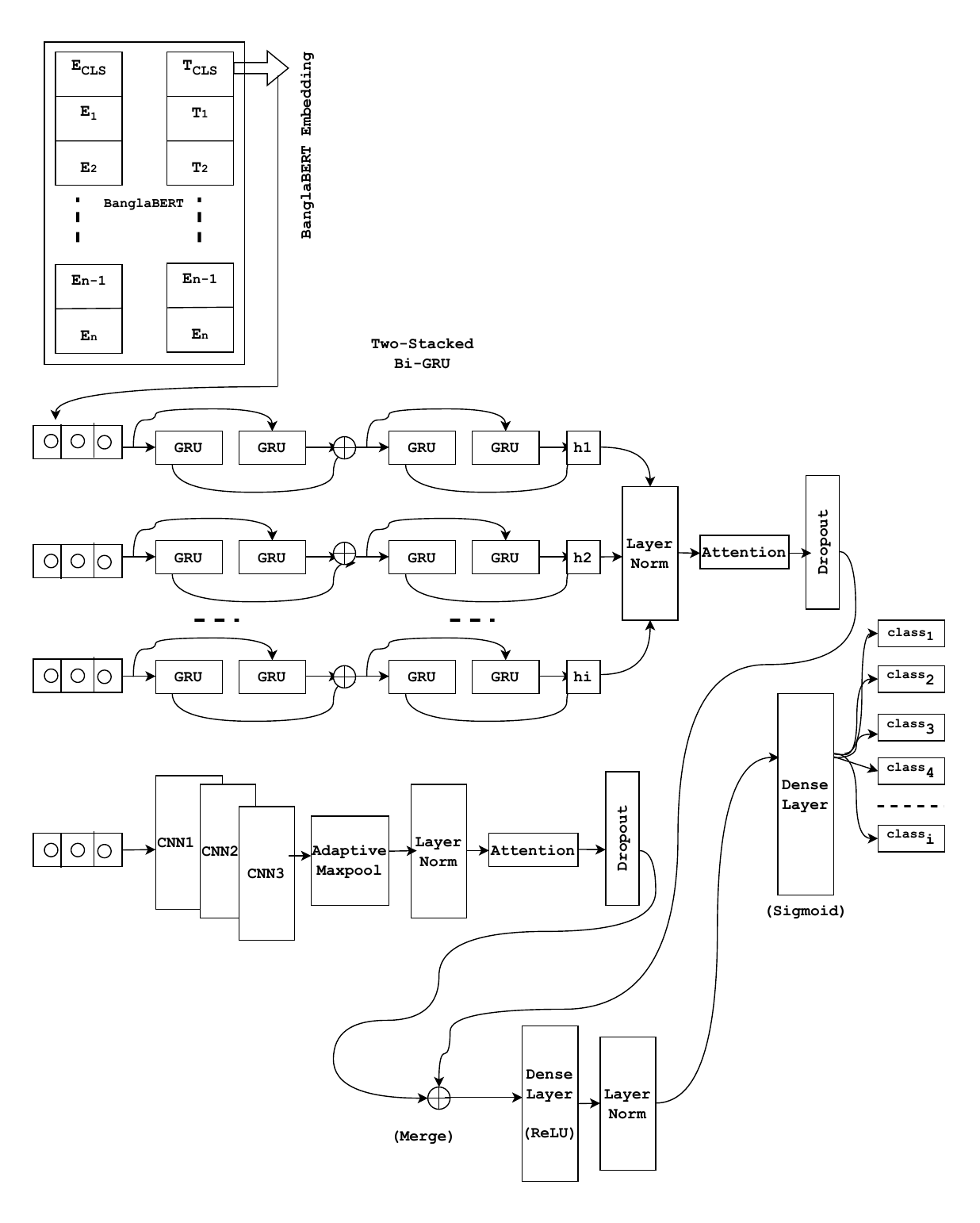}
    \caption{Proposed Model}
    \label{fig:proposed_model}
\end{figure*}

\section{System Overview}

The BanglaBERT model serves as the foundational component for generating contextual embeddings in the proposed architecture. It processes input sequences through transformer layers. This enables the capture of intricate linguistic nuances and contextual dependencies inherent in Bangla text, addressing language-specific biases observed in general-purpose models. The outputs are subsequently fed into parallel CNN and Bi-GRU branches for specialized feature extraction.

\subsection{Bi-GRU with Attention}
The bidirectional Gated Recurrent Unit\citep{cho-etal-2014-properties} is employed to model sequential dependencies within the BERT embeddings, enhancing the representation of temporal and contextual patterns in hate speech detection. Configured with two layers and a hidden dimension of 128, the Bi-GRU processes the input in both forward and backward directions. Layer normalization is then applied to stabilize training. Subsequently, a self-attention mechanism, with one head is applied to focus on salient sequential features.

\subsection{CNN with Attention}
The Convolutional Neural Network \citep{6795724} functions as a local feature extractor, capturing n-gram patterns from the BERT embeddings through parallel convolutions with kernel sizes of 1, 2, and 3 each producing 128 filters. Then ReLU activation is applied, followed by adaptive max pooling to aggregate the features. Layer normalization is then used to normalize the concatenated outputs. A self-attention layer is then utilized on the expanded CNN outputs to emphasize critical local patterns, generating attended features that complement the sequential modeling.

\subsection{Feature Fusion Layer}
The feature fusion layer integrates the outputs from the CNN and Bi-GRU branches to create a cohesive representation for classification. The CNN features and Bi-GRU features are concatenated which is then projected through a linear layer to a hidden dimension of 128, followed by ReLU activation and layer normalization.

\subsection{Output Layer}
The output layer comprises a final linear classifier that maps the fused features (dimension 128) to the number of target labels, producing logits for hate speech categories or target groups.

\section{Experimental Setup}

Table \ref{tab:hyper_para}, we illustrate the hyper-parameter setting of our proposed model. We utilized the Kaggle platform for experimental purposes and implemented our model using Pytorch.

\begin{table}[ht]
    \centering
    % \small % Reduce font size for readability
    \begin{tabular}{l *{2}{c}} % 1 left-aligned column (languages) + 4 centered columns (participants)
        \hline
        \textbf{Parameters} & \textbf{Value}\\
        \hline
        Batch size  & 16 \\
        Learning rate & $1 \times 10^{-5}$ \\
        Loss function & CrossEntropyLoss \\
        Optimizer & AdamW \\
        Dropout & 0.3 \\
        Hidden Units & 128 \\
        Max sequence length & 128 \\
        Gradient clipping & Yes \\
        \hline
    \end{tabular}
    \caption{Hyperparameter values}
    \label{tab:hyper_para}
\end{table}

\section{Results}

In Subtask 1A of BLP-2025 Task 1, we achieved 2nd place with micro F1-Score of 0.7345 (Table \ref{tab:BLP-2025 Task 1 (Subtask 1A)}), closely trailing the top-ranked team shifat\_islam (0.7362). The performance gap of only 0.0017 indicates high competitiveness and suggests that the proposed system is effective in handling the classification challenges presented in this subtask.

\begin{table}[ht]
    \centering
    \begin{tabular}{ccc}
        \hline
        \textbf{Teams} & \textbf{Rank} & \textbf{Score (F1-Micro)} \\
        \hline
        shifat\_islam & 1 & 0.7362 \\
        \textbf{SyntaxMind} & \textbf{2} & \textbf{0.7345} \\
         zannatul\_007 & 3 & 0.734 \\
        mahim\_ju &  4 & 0.7331 \\
         mohaymen &  15 & 0.7133 \\
         im\_tushu\_221 & 25 & 0.6901 \\
         Organizers &  35 & 0.5638 \\
         \hline
    \end{tabular}
    \caption{Performance Ranking of Participating Teams in BLP-2025 Task 1 (Subtask 1A)}
    \label{tab:BLP-2025 Task 1 (Subtask 1A)}
\end{table}

In Subtask 1B, our system secured the 5th position with a score of 0.7317 (Table \ref{tab:BLP-2025 Task 1 (Subtask 1B)}). Although the rank is slightly lower compared to Subtask 1A, the performance remains competitive, with a minimal margin separating the top five teams. The highest score in this subtask (0.7356 by mahim\_ju) exceeds our result by only 0.0039, indicating that the system maintains consistent performance across both subtasks.

\begin{table}[ht]
    \centering
    \begin{tabular}{ccc}
        \hline
        \textbf{Teams} & \textbf{Rank} & \textbf{Score (F1-Micro)} \\
        \hline
        mahim\_ju & 1 & 0.7356 \\
         shifat\_islam & 2 & 0.7335 \\
        mohaiminulhoque &  3 & 0.7328 \\
         reyazul &  4 & 0.7317 \\
         \textbf{SyntaxMind} & \textbf{5} & \textbf{0.7317} \\
        ashraf\_989 & 15 & 0.7114 \\
         Organizers & 23 & 0.5974 \\
         \hline
    \end{tabular}
    \caption{Performance Ranking of Participating Teams in BLP-2025 Task 1 (Subtask 1B)}
    \label{tab:BLP-2025 Task 1 (Subtask 1B)}
\end{table}

\section{Conclusion}

In this study, we presented our system developed for BLP-2025 Task 1, participating in both Subtask 1A and Subtask 1B. A single, unified model architecture was employed across both subtasks, demonstrating strong and consistent performance. The results highlight the effectiveness of our proposed hybrid ensemble approach in addressing Bangla hate speech detection. In future work, we aim to extend and adapt our system to Subtask 1C to further evaluate the performance on multiple task specific situation.

\bibliography{custom}

\end{document}